\documentclass[lettersize,journal]{IEEEtran}
\usepackage{amsmath,amsfonts}
\usepackage{algorithmic}
\usepackage{array}
\usepackage[caption=false,font=normalsize,labelfont=sf,textfont=sf]{subfig}
\usepackage{textcomp}
\usepackage{stfloats}
\usepackage{url}
\usepackage{verbatim}
\usepackage{graphicx}
\usepackage{multirow}
\usepackage[english]{babel}
\usepackage[utf8]{inputenc}
\usepackage{algorithm}
\usepackage{siunitx}
\usepackage{mathrsfs}
\usepackage{tikz}
\usetikzlibrary{shapes,arrows}
\usepackage{xcolor}
\usepackage[numbers]{natbib}
\def\BibTeX{{\rm B\kern-.05em{\sc i\kern-.025em b}\kern-.08em
    T\kern-.1667em\lower.7ex\hbox{E}\kern-.125emX}}
\usepackage{balance}
\usepackage{hyperref}
\begin{document}
\title{Enhanced Feature Based Granular Ball Twin Support Vector Machine}
\author{A. Quadir, M. Sajid, M. Tanveer{$^*$}, ~\IEEEmembership{Senior Member,~IEEE}, P. N. Suganthan,~\IEEEmembership{Fellow,~IEEE}, for the Alzheimer’s Disease Neuroimaging Initiative
\thanks{ \noindent $^*$Corresponding Author\\
 A. Quadir, M. Sajid, and M. Tanveer are with the Department of Mathematics, Indian Institute of Technology Indore, Simrol, Indore, 453552, India (e-mail: mscphd2207141002@iiti.ac.in, phd2101241003@iiti.ac.in and mtanveer@iiti.ac.in).\\
 Ponnuthurai N. Suganthan is with the KINDI Center for Computing Research, College of Engineering, Qatar University, Doha 2713, Qatar (e-mail: p.n.suganthan@qu.edu.qa) \\
 This study used data from the Alzheimer's Disease Neuroimaging Initiative (ADNI) (\url{adni.loni.usc.edu}). The ADNI investigators were responsible for the design and implementation of the dataset, but they did not take part in the analysis or the writing of this publication.  \url{http://adni.loni.usc.edu/wp-content/uploads/how\_to\_apply/ADNI\_Acknowledgement\_List.pdf} has a thorough list of ADNI investigators.
 }}
 
\maketitle
\begin{abstract}
In this paper, we propose enhanced feature based granular ball twin support vector machine (EF-GBTSVM). EF-GBTSVM employs the coarse granularity of granular balls (GBs) as input rather than individual data samples. The GBs are mapped to the feature space of the hidden layer using random projection followed by the utilization of a non-linear activation function. The concatenation of original and hidden features derived from the centers of GBs gives rise to an enhanced feature space, commonly referred to as the random vector functional link (RVFL) space. This space encapsulates nuanced feature information to GBs. Further, we employ twin support vector machine (TSVM) in the RVFL space for classification. TSVM generates the two non-parallel hyperplanes in the enhanced feature space, which improves the generalization performance of the proposed EF-GBTSVM model. Moreover, the coarser granularity of the GBs enables the proposed EF-GBTSVM model to exhibit robustness to resampling, showcasing reduced susceptibility to the impact of noise and outliers. We undertake a thorough evaluation of the proposed EF-GBTSVM model on benchmark UCI and KEEL datasets. This evaluation encompasses scenarios with and without the inclusion of label noise. Moreover, experiments using NDC datasets further emphasize the proposed model’s ability to handle large datasets. Experimental results, supported by thorough statistical analyses, demonstrate that the proposed EF-GBTSVM model significantly outperforms the baseline models in terms of generalization capabilities, scalability, and robustness. Furthermore, we applied our proposed EF-GBTSVM model to the Schizophrenia and Alzheimer’s Disease Neuroimaging Initiative (ADNI) datasets, demonstrating the models' effectiveness in real-world applications. 
\end{abstract}
\section{Introduction}
\IEEEPARstart{S}{upport} vector machine (SVM) \cite{cortes1995support} stands out as one of the most extensively employed machine learning (ML) models for classification problems. The main objective of SVM is to find the optimal hyperplane, effectively segregating the classes while simultaneously maximizing the margin between the two classes. SVM has found its applications in various real-world problems such as cancer detection \cite{akhtar2023roboss}, brain-computer interface \cite{molina2003joint}, diagnosis of 
Schizophrenia disease \cite{quadir2024granularpin}, significant memory concern diagnosis \cite{sajid2024decoding}, and so on. 
\par
SVM addresses a single large quadratic programming problem (QPP), which increases computational complexity and makes it less effective for handling large-scale datasets. To address the computational complexity of SVM, \citet{khemchandani2007twin} introduced twin SVM (TSVM). TSVM solves two smaller-sized QPPs instead of a single large QPP, making TSVM four times faster compared to SVM \cite{khemchandani2007twin,tanveer2022comprehensive}. This firmly establishes TSVM as a standout and superior choice due to its efficiency. However, TSVM still requires the computation of matrix inverses and relies on the existence of nonsingular matrices making it not suitable for large scale problems. Numerous adaptations of the TSVM have been suggested to tackle various challenges encountered in classification problems \cite{akhtar2024advancing, quadir2024multiview, quadir2024enhancing, quadir2024intuitionisticUniversum}. Notably, among these challenges, the presence of noise stands out as a significant issue. To mitigate the impact of noise, diverse techniques are employed to assign fuzzy membership weights to noisy data points across various applications \cite{rezvani2019intuitionistic, quadir2024intuitionistic}. Recently, the granular ball SVM (GBSVM) \cite{xia2022gbsvm} has been proposed, which integrates the concepts of SVM with granular computing. GBSVM addresses a single quadratic programming problem using the PSO algorithm, which can sometimes lead to convergence at local minima. To overcome this limitation, granular ball TSVM (GBTSVM) and its large-scale variant is proposed in \cite{quadir2024granularTSVM, quadir2024granularpin}. These models solve two complex quadratic programming problems, improving the performance and robustness of the model. GBTSVM exhibits good performance in effectively managing datasets that are contaminated with noise and outliers.
Artificial neural networks (ANNs) are ML models that mimic the structure and functionality of the human brain's neural system. Within ANNs, nodes, also referred to as ``neurons,'' are interconnected in layers. These layers collaborate to process, analyze, and relay information, ultimately enabling the network to make predictions or decisions. ANN has showcased achievements across diverse fields, including stock market prognostication \cite{dase2010application}, rainfall forecasting \cite{luk2001application}, clinical medicine \cite{baxt1995application}, solving partial and ordinary differential equations \cite{lagaris1998artificial,lagaris2000neural}, diagnosis of Alzheimer's disease \cite{sajid2024intuitionistic}, feature interpretability \cite{sajid2024neuro} and so on. In addition to numerous advantages, there are certain drawbacks associated with ANN models, such as slow convergence, local minima problems, and sensitivity to learning rates. 

To address these challenges, randomized neural networks (RNNs) based on closed-form solutions \cite{suganthan2018non} have been proposed. Generally, a certain level of randomness is inherent in either the structure or the learning process of the RNN model. The presence of randomness in RNN provides them with the capability to learn with fewer tunable parameters in a shorter duration, often eliminating the need for advanced hardware. The random vector functional link (RVFL) neural network \cite{pao1994learning,malik2023random} is a widely recognized variant of RNNs. In the RVFL framework, the weights linking the input layer to the hidden layer are randomly generated from a pertinent domain and remain constant throughout the training phase. The direct connections within the RVFL play a pivotal role in determining its generalization performance \cite{zhang2016comprehensive, vukovic2018comprehensive}. The output parameters, encompassing the weights of direct links and the connections linking the hidden layer to the output layer, are analytically computed using techniques such as the least squares method or the pseudo-inverse. Furthermore, the RVFL's thinner topology, when contrasted with the other popular RNNs such as extreme learning machine or RVFL without direct link (RVFLwoDL) \cite{huang2006extreme}, contributes to reduced complexity, aligning with the probably approximately correct (PAC) learning theory and Occam's principle \cite{shi2021random}. The RVFL provides rapid training speed while also possessing universal approximation capabilities \cite{igelnik1995stochastic,needell2020random}. The RVFL demonstrates promising results across various applications, such as data streams \cite{pratama2018parsimonious}, DNA-binding proteins prediction \cite{quadir2024multiviewrandom}, Alzheimer's disease diagnosis \cite{tanveer2024fuzzy, tanveer2024ensemble}, and so on. 

Input data samples offer a wide range of information derived from various feature representations. This includes compressed feature representations obtained from lower-dimensional feature spaces as well as sparse feature representations derived from higher-dimensional feature spaces \cite{hinton2006fast,hinton2006reducing}. Different learning algorithms explore various underlying information present in the data through these distinct feature representations. RVFL utilizes random feature transformation in conjunction with the original features and has been effectively employed in tasks related to both classification and regression. Inspired by the achievements associated with diverse feature representations, we propose an enhanced feature-based granular ball twin support vector machine (EF-GBTSVM). EF-GBTSVM model first generates the GBs from the input training data. The centers of generated GBs are then projected to the hidden layer and weights and biases are generated randomly. The hidden layer, endowed with an activation function, serves to convert the input feature space into a randomized feature space. TSVM model is used to train for classification over the enhanced feature space, which is a combination of original features and hidden features of GB centers. The proposed EF-GBTSVM utilizes GBs as input for classifier construction, ensuring heightened robustness, resilience to resampling, and computational efficiency. \\
The proposed EF-GBTSVM model exhibits several notable characteristics:
\begin{enumerate}
    \item EF-GBTSVM utilizes GBs as inputs and constructs the classifier in the RVFL space, providing enhanced robustness, resilience to resampling, and computational efficiency. 
    \item Leveraging the principles of granularity, the proposed EF-GBTSVM model effectively addresses the negative impacts of noise and outliers. 
    \item Training the proposed EF-GBTSVM model in the enhanced feature space elevates its performance by effectively capturing intricate data patterns and complex relationships.
    \item The proposed EF-GBTSVM model achieves scalability by using a coarser granularity, allowing it to efficiently manage large datasets.
    \item To illustrate the practical applicability of the proposed EF-GBTSVM model, we apply them to real-world datasets, specifically the Schizophrenia (SCZ) dataset and the ADNI datasets for classifying SCZ and Alzheimer’s disease (AD).
\end{enumerate} 
The rest of the paper is organized as follows: Section \ref{Related work} gives the information of the related work. We discuss the proposed model in Section \ref{proposed model}. Section \ref{sec4} gives a theoretical comparison of the proposed EF-GBTSVM model and the baseline models. The scalability of the proposed model is also shown in terms of time complexity in Section \ref{sec5}. Section \ref{Experimental Results} demonstrates the experimental results. Finally, in Section \ref{Conclusions} we conclude by suggesting potential directions for future research. 
\section{Related Work}
\label{Related work}
This section provides a concise overview of granular ball computing and TSVM. 
\subsection{Notations}
Let $\mathcal{X} =\{(x_i, y_i), i=1,2,3, \ldots, n\}$ denotes the training dataset, where $y_i \in \{+1,-1\} $ represents the label of $x_i \in \mathbb{R}^{1 \times m}$. The collection of generated granular balls is represented as $G=\{GB_i, \hspace{0.2cm} i=1,2, \ldots, k\} = \{(c_i, t_i),\hspace{0.2cm} i=1,2, \ldots, k\}$, where $c_i$ signifies the center, and $t_i$ is the label of the $i^{th}$ granular ball. Let $X=(x_1^t, x_2^t, \ldots, x_n^t)^t$ represent the collection of all input samples, where $(\cdot)^{t}$ represent the transpose operator.


\subsection{Granular Ball Computing}
In 1996, Lin and Zadeh proposed the idea of ``granular computing'' aiming to reduce the number of required training data points \cite{xia2019granular}. The fundamental concept of granular ball computing involves using a hyper ball to enclose either the entire sample space or a specific portion of it. Using the ``granular ball'' to represent the sample space helps capture multi-granularity learning attributes and allows for a more precise characterization of the sample space. The center ``$c$'' of a \( GB \) is defined as the centroid calculated from all sample points within the ball. Mathematically, it can be expressed as: $c=\frac{1}{l}\sum_{i=1}^lx_i$, where $x_i$ signifies an individual data point and $l$ represents the total number of data points contained within the granular ball. The label assigned to the granular ball is selected by identifying the label that occurs most frequently among the samples contained within the granular ball. To measure the degree of division within a granular ball, the concept of ``threshold purity'' is introduced. This threshold refers to the proportion of samples within a granular ball that possesses identical labels, particularly the predominant labels.

Let $\mathcal{X}=\{(x_i, \hspace{0.1cm} y_i), i=1,2,3, \ldots, n\}$ be the training dataset. The generated granular ball from the dataset $\mathcal{X}$ is represented as $GB_j$ $(j =1, 2, 3, \ldots, k)$. Here $k$ denotes the total number of granular balls generated from the dataset $\mathcal{X}$. Formally, the solution for generating granular balls is defined by the following optimization problem:
\begin{align}
    & min \hspace{0.2cm} \vartheta_1 \times \frac{n}{\sum_{i=1}^{k}\lvert GB_i \rvert} + \vartheta_2 \times k \nonumber \\
    s.t. & \hspace{0.2cm} quality(GB_j) \geq \varphi, \hspace{0.2cm} j=1,2, \ldots, k,
\end{align}
where $\varphi$ denotes the purity threshold, while $\vartheta_1$ and $\vartheta_2$ denote the weight coefficients.
\subsection{Twin Support Vector Machine (TSVM)}
The main idea of twin support vector machine (TSVM) \cite{khemchandani2007twin} is to generate two non-parallel hyperplanes, with each plane passing through the corresponding samples of the respective classes and maximizing the distance of the hyperplanes from samples of the other class. Let $A \in \mathbb{R}^{n_1 \times m}$ and $B \in \mathbb{R}^{n_2 \times m}$ are the input matrices, where $n_1$ ($n_2$) is the number of data samples belonging to $+1$ ($-1$) class and $m$ is the total number of attributes of each data sample. The formulation of TSVM is given as follows:
\begin{align}
      & \underset{w_1,b_1}{min} \hspace{0.2cm} \frac{1}{2} \|Aw_1 + e_1b_1\|^2 + d_1e_2^t\xi_2 \nonumber \\
     & s.t. \hspace{0.2cm} -(Bw_1 + e_2b_1) + \xi_2 \geq e_2, \nonumber \\
     & \hspace{0.8cm} \xi_2 \geq 0,
\end{align}
and
\begin{align}
    & \underset{w_2,b_2}{min} \hspace{0.2cm} \frac{1}{2} \|Bw_2 + e_2b_2\|^2 + d_2e_1^t\xi_1 \nonumber \\
     & s.t. \hspace{0.2cm} (Aw_2 + e_1b_2) + \xi_1 \geq e_1, \nonumber \\
     & \hspace{0.8cm} \xi_1 \geq 0,
\end{align}
here $\xi_1$ and $\xi_2$ denotes the slack vectors, $d_1$ and $d_2$ represent the pre-specified penalty parameters and $e_1$ and $e_2$ are vectors composed of ones with approximate dimensions.
\section{Proposed Enhanced Feature Based Granular Ball Twin Support Vector Machine}
\label{proposed model}
In this section, we propose enhanced feature-based granular ball twin support vector machine (EF-GBTSVM). EF-GBTSVM model utilizes granular balls (GBs) as inputs and constructs the classifier within the RVFL space, offering enhanced robustness, resilience to resampling, and computational efficiency. By leveraging the principles of granularity, the proposed EF-GBTSVM model effectively mitigates the negative impacts of noise and outliers, thus addressing the underlying problem of classification in the presence of noisy data. The proposed model can be explained through a three-step process: the initial step involves generating the GB from the input training data. The second step involves feature mapping, where the features of the center of GBs are transformed into an enhanced feature representation (see eqns. \ref{eq:RVFL_1} and \ref{eq:RVFL_2}). The third step is to construct a classifier using TSVM over the enhanced feature space. TSVM determines the non-parallel decision hyperplanes by utilizing the enhanced features of centers of GBs rather than the original data points.  

Let $C \in \mathbb{R}^{k \times m}$ be the centers of the generated GBs of the training dataset $\mathcal{X}$. The matrix \( C \) of the GB is processed through the hidden layers to extract the more important features.
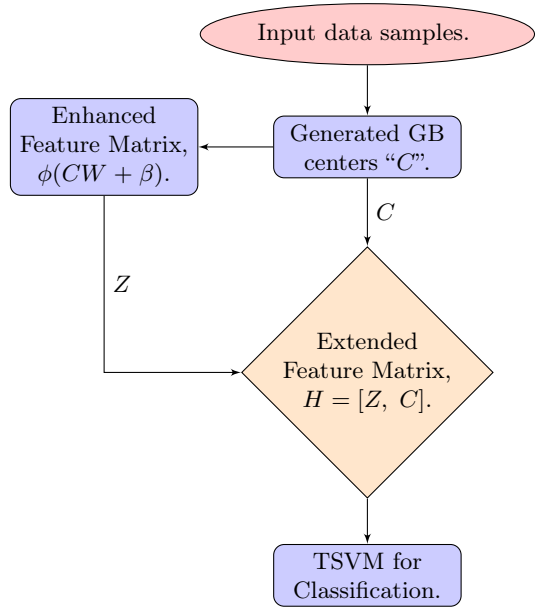
\begin{figure}[ht]
    \centering       
\tikzstyle{decision} = [diamond, draw, fill=orange!20, 
    text width=7.0em, text badly centered, node distance=0.5cm, inner sep=0pt]
\tikzstyle{block} = [rectangle, draw, fill=blue!20, 
    text width=7em, text centered, rounded corners, minimum height=2.5em]
\tikzstyle{line} = [draw, -latex']
\tikzstyle{cloud} = [draw, ellipse,fill=red!20, node distance=1cm,
    minimum height=2.5em,text width=9em]
\begin{tikzpicture}[node distance = 1.5cm, auto]
    \node [cloud] (init) {Input data samples.};
    \node [block, below of=init] (identify) {Generated GB centers ``$C$''.};
    \node [block, left of=identify, node distance=3.5cm] (update) {Enhanced Feature Matrix, $\phi(CW + \beta)$.};
    \node [decision, below of=identify,node distance=3cm] (decide) {Extended Feature Matrix, $H=[Z, \hspace{0.1cm} C]$.};
    \node [block, below of=decide, node distance=2.7cm] (stop) {TSVM for Classification.};
    \path [line] (init) -- (identify);
    \path [line] (identify) -- node {$C$} (decide);
    \path [line] (update) |- node [near start] {$Z$} (decide);
    \path [line] (identify) -- (update);
    \path [line] (decide) -- (stop);
\end{tikzpicture}
\caption{Flowchart of the proposed EF-GBTSVM model. The entire dataset can be considered as a granular ball (GB). First, calculate the center ``$C$'' and the label of the GBs. Then, compute the hidden layer matrix ``$Z$'' for the generated GB center, with weights and biases randomly initialized. Next, obtain the enhanced features (RVFL features) by concatenating the hidden feature ``$Z$'' with the center matrix ``$C$''. Finally, use TSVM to classify the data points into $+1$ and $-1$ classes, respectively.}
    \label{Proposed RV-GBTSVM model architecture}
\end{figure}
Consider a hidden layer with $h$ nodes, $W \in \mathbb{R}^{m \times h} $ represents the weight matrix from the input layer (of centers) to the hidden layer, and $\beta$ denotes the bias term. The hidden layer matrix $Z$ is defined as:
\begin{align}
\label{eq:RVFL_1}
    Z=\phi(CW + \beta) \in \mathbb{R}^{k \times h}, 
\end{align}
where $\phi$ is the activation function. The enhanced features (RVFL features) are obtained by concatenating the hidden feature $Z$ and the center $C$ of generated GB of training dataset and is defined as:
\begin{align}
\label{eq:RVFL_2}
    H = [Z, \hspace{0.2cm} C] \in \mathbb{R}^{k \times (m+h)}.
\end{align}
Now, we train our proposed model on enhanced feature space rather than the original feature space. This novel training approach enables the proposed model to capture nuanced features that traditional models like TSVM often overlook. Fig. \ref{Proposed RV-GBTSVM model architecture} depicts the flowchart of the proposed EF-GBTSVM model and illustrates the process of transforming the entire dataset into GB. It shows the calculation of the center ``$C$'' and the hidden layer matrix ``$Z$''. The enhanced features are obtained by concatenating the hidden feature ``$Z$'' with the center matrix ``$C$'', followed by the TSVM classification into $+1$ and $-1$ classes.

Let $H=T_1 \cup T_2$, where $T_1 = [Z^+, \hspace{0.2cm} C^+] \in \mathbb{R}^{k_1 \times (m+h)}$ and $T_2 = [Z^-, \hspace{0.2cm} C^-] \in \mathbb{R}^{k_2 \times (m+h)} $ represent the enhanced feature matrices of $+1$ and $-1$ class. Here, $Z^+$ and $Z^-$ represent the hidden layer matrix $+1$ and $-1$ class and $C^+$ and $C^-$ represent the generated GB center matrix of $+1$ and $-1$ class, and $k_1$ and $k_2$ are the number of GBs of $+1$ and $-1$ class, respectively. The objective function of the proposed EF-GBTSVM model is formally defined as follows:
\begin{align}
\label{eq:6}
      & \underset{w_1,b_1}{min} \hspace{0.2cm} \frac{1}{2} \|T_1w_1 + e_1b_1\|^2 + d_1e_2^t\xi_2 \nonumber \\
     & s.t. \hspace{0.2cm} -(T_2w_1 + e_2b_1) + \xi_2 \geq e_2, \nonumber \\
     & \hspace{0.8cm} \xi_2 \geq 0,
\end{align}
and
\begin{align}
\label{eq:7}
    & \underset{w_2,b_2}{min} \hspace{0.2cm} \frac{1}{2} \|T_2w_2 + e_2b_2\|^2 + d_2e_1^t\xi_1 \nonumber \\
     & s.t. \hspace{0.2cm} T_1w_2 + e_1b_2 + \xi_1 \geq e_1, \nonumber \\
     & \hspace{0.8cm} \xi_1 \geq 0,
\end{align}
where $w_1 \in \mathbb{R}^{(m+h)\times 1}$, $b_1 \in \mathbb{R}$, $w_2 \in \mathbb{R}^{(m+h)\times 1}$, $b_2 \in \mathbb{R}$, $\xi_2 \in \mathbb{R}^{k_2 \times 1}$, and $\xi_1 \in \mathbb{R}^{k_1 \times 1}$. $d_1$ and $d_2$ are the tunable parameters. $e_1$ and $e_2$ are the vectors of ones of appropriate dimensions. 

The Lagrangian corresponding to the problem \eqref{eq:6} is given by
\begin{align}
\label{eq:8}
    L&=\frac{1}{2}\|T_1w_1 + e_1b_1\|^2 + d_1e_2^t\xi_2 - \alpha^t(-(T_2w_1 + e_2b_1) \nonumber \\
    & + \xi_2 - e_2) - \beta^t\xi_2,
\end{align}
where $\beta \in \mathbb{R}^{k_1 \times 1}$ and $\alpha \in \mathbb{R}^{k_2 \times 1}$ are the vectors of Lagrangian multipliers.\\
The optimal conditions are given as follows:
\begin{align}
    &T_1^t(T_1w_1 + e_1b_1) + T_2^t\alpha = 0, \label{eq:10}\\
    &e_1^t(T_1w_1 + e_1b_1) + e_2^t\alpha = 0, \label{eq:11}\\
    &e_2d_1 - \alpha - \beta =0, \label{eq:12}\\
    &-(T_2w_1 + e_2b_1) + \xi_2 \geq e_2, \hspace{0.3cm} \xi_2 \geq 0, \\
    &\alpha^t(-(T_2w_1 + e_2b_1) + \xi_2 - e_2) = 0, \hspace{0.3cm} \beta^t\xi_2 = 0,\\
    &\alpha \geq 0, \hspace{0.3cm} \beta \geq 0.
\end{align}
Combining \eqref{eq:10} and \eqref{eq:11} leads to
\begin{align}
\label{eq:16}
    \binom{T_1^t}{e_1^t} \left(T_1 \hspace{0.2cm} e_1\right)\binom{w_1}{b_1} + \binom{T_2^t}{e_2^t}\alpha = 0.
\end{align}
Let $H=\left(T_1 \hspace{0.2cm} e_1\right),$ $G=\left(T_2 \hspace{0.2cm} e_2\right)$ and $u_1=\binom{w_1}{b_1}$ then, \eqref{eq:16} can be reformulated as:
\begin{align}
    &H^tHu + G^t\alpha = 0, \nonumber \\
    i.e., \hspace{0.2cm} & u_1 = -(H^tH)^{-1}G^t\alpha \label{eq:17}.
\end{align}
Computing the inverse of $H^tH$ presents a formidable challenge. However, this difficulty can be effectively addressed by incorporating a regularization term denoted as $\delta I$ in \eqref{eq:17}, where $I$ represents an identity matrix of suitable dimensions. Thus,
\begin{align}
\label{eq:18}
    u_1 = -(H^tH+\delta I)^{-1}G^t\alpha.
\end{align}
Using eqn. \eqref{eq:18} and the above K.K.T. conditions, we can obtain the dual of \eqref{eq:6} as follows:
\begin{align}
\label{eq:19}
     \underset{\alpha}{max} & \hspace{0.2cm} \alpha^te_2 - \frac{1}{2} \alpha^t E(F^tF + \delta I)^{-1}E^t \alpha \nonumber \\
     s.t. & \hspace{0.2cm} 0 \leq \alpha \leq d_1e_2,
\end{align}
Likewise, the Wolfe dual for \eqref{eq:7} can be obtained as
\begin{align}
\label{eq:13}
     \underset{\gamma}{max} & \hspace{0.2cm} \gamma^te_1 - \frac{1}{2} \gamma^t F(E^tE + \delta I)^{-1}F^t \gamma \nonumber \\
     s.t. & \hspace{0.2cm} 0 \leq \gamma \leq d_2e_1, 
\end{align}
where $F=\left(T_1 \hspace{0.2cm} e_1\right),$ $E=\left(T_2 \hspace{0.2cm} e_2\right)$ and $u_1=\binom{w_1}{b_1}$, $u_2=\binom{w_2}{b_2}$ is calculated as:
\begin{align}
\label{eq:14}
    u_1 = -(F^tF+\delta I)^{-1}E^t\alpha \hspace{0.1cm} \text{and} \hspace{0.1cm} u_2=(E^tE+\delta I)^{-1}F^t\gamma.
\end{align}
Analogously, $u_2=\binom{w_2}{b_2}$ corresponding to the $-1$ class can be calculated by the subsequent eqn.:
\begin{align}
\label{eq:21}
    u_2=(G^tG+\delta I)^{-1}H^t\gamma.
\end{align}
Once the optimal $u_1$ and $u_2$ are obtained, the following decision function can be used to predict the target value of a new sample:
\begin{align}
\label{eq:22}
    class(x) =  \underset{i \in \{1, 2\}}{\arg\min} \frac{\lvert w_i^tx + b_i\rvert}{\|w_i\|}.
\end{align}

\section{Comparison of the proposed EF-GBTSVM model w.r.t. the baseline GBSVM and TSVM models} \label{sec4}
This section outlines the difference between the proposed EF-GBTSVM model and existing GBSVM and TSVM models.
\begin{itemize}
        \item EF-GBTSVM vs GBSVM
            \begin{itemize}
                \item The proposed EF-GBTSVM model solves two quadratic programming problems (QPPs) to determine the optimal parameters. However, GBSVM solves one large QPP to obtain the optimal hyperplanes, leading to an increase in time complexity compared to the proposed models.
            \item The proposed models utilize the external package ``CVXOPT" to solve the dual of the QPPs, employing the ``qp-solvers" function to obtain the global solution, whereas, GBSVM employs the PSO algorithm (an iterative method), which may converge to local minima rather than the global minimum.
            \end{itemize}
         \item EF-GBTSVM vs TSVM
            \begin{itemize}
            \item The proposed model's effectiveness is attributed to its utilization of granular balls as inputs rather than individual sample points. This allows the EF-GBTSVM to efficiently handle large datasets and demonstrate scalability. However, the TSVM's imperative demand for matrix inversions presents formidable obstacles to its efficiency and applicability on large-scale datasets.
            \item  TSVM struggles to handle noise and outliers in datasets, whereas the proposed model addresses these challenges effectively by incorporating granular balls to generate optimal classifiers.
        \end{itemize}
\end{itemize}

\section{Time complexity and algorithm of the proposed EF-GBTSVM model} \label{sec5}
The complexity of the proposed EF-GBTSVM model mainly hinges on three factors: (a) the computation of granular balls, (b) the necessity of matrix multiplication to generate the hidden feature matrix, and (c) the use of TSVM for classification. The time complexity of standard TSVM \cite{khemchandani2007twin} is $\mathcal{O}(\frac{n^3}{4})$. Our approach begins with the training dataset \( \mathcal{X} \), which we consider as the initial granular ball (GB) set. We initially divide this GB into two granular balls using the 2-means clustering method, with a time complexity of \(\mathcal{O}(2n)\). In subsequent phases, if both granular balls are impure, they are further divided into four granular balls, each maintaining a maximum time complexity of \(\mathcal{O}(2n)\). This iterative process continues for a total of \( iter \) iterations. Therefore, the overall time complexity of generating granular balls is \(\mathcal{O}(iter \times 2n)\) or less, accounting for the maximum time complexity per iteration and the total number of iterations \( iter \). Generating the hidden feature matrix \( H \) involves multiplying the generated granular ball center matrix by randomly generated weights, with a time complexity of \( \mathcal{O}(k^2h) \). Hence, the overall time complexity of the proposed EF-GBTSVM is (or less than) $\mathcal{O}(\frac{k^3}{4}) + \mathcal{O}(iter \times 2n) + \mathcal{O}(k^2h)$, where $k$ represents the total number of generated granular balls and $h$ represents the number of hidden nodes. The detailed algorithm of the proposed EF-GBTSVM model, as outlined in \ref{Algorithm_EF-GBTSVM.}.

\begin{algorithm}[ht!]
\caption{Algorithm of the proposed EF-GBTSVM model.}
\label{Algorithm_EF-GBTSVM.}
\textbf{Input:} Purity threshold $\eta$, and the traning dataset $\mathcal{X}$. \\
\textbf{Output:} Model parameters.\\ \vspace{-5mm}
\begin{algorithmic}[1]
\STATE Assume the entire dataset $\mathcal{X}$ is represented as a granular ball $GB$ and set of granular balls, $G$, to be empty set, i.e., $GB=T$ and $G=\{\hspace{0.1cm}\}$. 
\STATE $Temp =\{GB\}$.
\FOR{$i = 1:\lvert Temp \rvert$}
\IF{$pur(GB_i) < \eta$}
\STATE Split $GB_i$ into $GB_{i1}$ and $GB_{i2}$, using $2$-means clustering algorithm. 
\STATE $Temp \leftarrow GB_{i1}, \hspace{0.05cm} GB_{i2}$. 
\ELSE 
\STATE Compute the center $c_i = \frac{1}{p} \sum_{j=1}^{p} x_j$ of $GB_i$, where $x_j \in GB_i$, $j=1, 2, \ldots, p$, and $p$ is the number of training sample in $GB_i$.  
\STATE Compute the label $t_i$ of $GB_i$, where $t_i$ is assigned the label of majority class samples within $GB_i$.
\STATE Put $GB_i = \{(c_i,t_i)\}$ in $G$. 
\ENDIF 
\ENDFOR  
\IF{$Temp \neq \{\hspace{0.1cm}\}$} 
\STATE Go to step 3 (for further splitting). 
\ENDIF
\STATE Set $G=\{GB_i, \hspace{0.2cm} i=1,2, \ldots, k\} = \{(c_i, t_i),\hspace{0.2cm} i=1,2, \ldots, k\},$ where $c_i$ signifies the center of the granular ball, $t_i$ is the label of $GB_i$ and $k$ is the number of generated granular balls. 
\STATE Find the hidden layer features using \eqref{eq:RVFL_1}.
\STATE Create the enhanced features using \eqref{eq:RVFL_2}. 
\STATE Compute $w_1$, $b_1$, $w_2$, and $b_2$ using \eqref{eq:19} and \eqref{eq:13}.
\STATE Classify testing samples into class $+1$ or $-1$ using \eqref{eq:22}.
\end{algorithmic}
\end{algorithm}

\section{Experimental Results}
\label{Experimental Results}
To evaluate the effectiveness of the proposed EF-GBTSVM model, we conduct a comparative analysis along with the baseline models on benchmark datasets from the UCI \cite{dua2017uci} and KEEL \cite{derrac2015keel} repository. Moreover, we performed experiments using datasets generated through the NDC Data Generator \cite{ndc}. Furthermore, we implement the proposed model on the Alzheimer’s dataset available through the Alzheimer’s Disease Neuroimaging Initiative (ADNI) and the Schizophrenia dataset provided by the Center for Biomedical Research Excellence (COBRE).
\subsection{Experimental Setup}
The experimental setup includes a PC with an Intel(R) Xeon(R) Gold $6226R$ CPU operating at $2.90$ GHz and $128$ GB of RAM. This system operates on the Windows 11 platform and utilizes Python $3.11$ for execution. The dual QPP arising in the proposed EF-GBTSVM model and the baseline models are solved by the ``qp-solvers'' function available in the CVXOPT package. The dataset is randomly partitioned into training and testing subsets at a ratio of $70:30$, respectively. We use 5-fold cross-validation method combined with a grid search approach to fine-tune the models' hyperparameters within designated ranges, $d_1 = d_2 = \{10^{-5}, 10^{-4}, \ldots, 10^{5}\}$. The number of hidden nodes is chosen from the range $3$ to $203$ with a step size of $20$. We tuned nine different activation functions. The indexing of these functions is as follows: 1) SELU, 2) ReLU, 3) Sigmoid, 4) Sine, 5) Hardlim, 6) Tribas, 7) Radbas, 8) Sign, and 9) Leaky ReLU.
\subsection{Experiments on Benchmark UCI and KEEL Datasets}
In this subsection, we analyze and compare the proposed EF-GBTSVM model along with the baseline SVM \cite{cortes1995support}, GBSVM \cite{xia2022gbsvm}, TSVM \cite{khemchandani2007twin}, RVFLwoDL \cite{huang2006extreme}, RVFL \cite{pao1994learning}, and GBTSVM \cite{quadir2024granularTSVM} models on $29$ UCI and KEEL benchmark datasets. Additionally, we apply the TSVM model on the hidden feature space of the GB centers to assess the significance of enhanced features compared to individually considering the hidden and original features - the resultant model is named hidden feature based granular ball TSVM (HF-GBTSVM). 

The experimental outcomes depicted in Table \ref{Classification performance in Linear Case.} represent the performance of the proposed models along with the baseline models. The accuracy-based comparison reveals that our proposed EF-GBTSVM demonstrates superior performance compared to the baseline SVM, GBSVM, TSVM, RVFLwoDL, RVFL, and GBTSVM models across the majority of datasets. From Table \ref{Classification performance in Linear Case.}, the average accuracy (ACC) for our proposed EF-GBTSVM model stands at $86.18\%$. In contrast, the average ACC of the SVM, GBSVM, TSVM, RVFLwoDL, RVFL, GBTSVM, and HF-GBTSVM models are $80.71\%$, $72.85\%$, $65.01\%$, $83.22\%$,  $84.02\%$, $84.23\%$ and $82.13\%$, respectively. The average ACC metric can be influenced by outstanding performance in a single dataset, which could compensate for weaker results across various datasets, potentially resulting in a biased measure. Therefore, we apply a ranking method to assess the effectiveness and evaluate the performance of the models. In this approach, classifiers are ranked based on their performance: models with better performance receive a lower rank, while those with poorer performance are given a higher rank. To evaluate $q$ models across $P$ datasets, \( r_j^i \) denotes the rank of the \( j^{th} \) model on the \( i^{th} \) dataset. $\mathscr{R}_j = \frac{1}{P}\sum_{i=1}^P r_j^i$ is the average rank of the model. The average rank of proposed EF-GBTSVM models along with the HF-GBTSVM, SVM, GBSVM, TSVM, RVFLwoDL, RVFL, and GBTSVM models are $2.42$, $4.42$, $4.97$, $6.02$, $7.55$, $3.85$, $3.76$, and $3.02$ respectively. The proposed EF-GBTSVM exhibits the most favorable average rank among the compared models. Therefore, the proposed EF-GBTSVM model demonstrates superior generalization ability compared to the baseline models. 
A notable finding is that the proposed EF-GBTSVM (on enhanced feature space) outperforms  HF-GBTSVM (on hidden feature space) and TSVM (on original feature space). This shows the clear significance of training the classifier over the enhanced feature space over the individual original and hidden feature spaces.
\begin{table*}[ht!]
\centering
    \caption{Comparison results on benchmark UCI and KEEL datasets for the proposed EF-GBTSVM model with the baseline models.}
    \label{Classification performance in Linear Case.}
    \resizebox{0.9\linewidth}{!}{
\begin{tabular}{lcccccccc}
\hline
Datasets                         & SVM \cite{cortes1995support}  & GBSVM \cite{xia2022gbsvm} & TSVM  \cite{khemchandani2007twin} & RVFLwoDL \cite{huang2006extreme}  & RVFL \cite{pao1994learning} & GBTSVM \cite{quadir2024granularTSVM} & HF-GBTSVM & EF-GBTSVM$^{\dagger}$ \\ \hline
aus & 88.46 & 81.11 & 64.31 & 87.98 & 86.94 & 89.06 & 85.1 & 88.96 \\
breast\_cancer & 72.09 & 62.79 & 60 & 74.42 & 72.09 & 73.26 & 73.26 & 78.14 \\
breast\_cancer\_wisc\_diag & 78.42 & 79.69 & 78.89 & 79.25 & 80.28 & 80.56 & 78.42 & 83.62 \\
checkerboard\_Data & 85.46 & 71.06 & 64.31 & 85.98 & 85.94 & 86.06 & 85.1 & 86.21 \\
chess\_krvkp & 84.67 & 69.62 & 67.41 & 90.2 & 90.41 & 85.19 & 90.09 & 98.16 \\
crossplane130 & 97.24 & 100 & 71.35 & 97.44 & 97.44 & 97.44 & 100 & 100 \\
ecoli-0-1\_vs\_2-3-5 & 81.89 & 77.3 & 68.24 & 80.59 & 80.59 & 85.14 & 90.54 & 82.43 \\
ecoli-0-1-4-6\_vs\_5 & 95.81 & 94.05 & 67.88 & 98.81 & 98.81 & 96.05 & 89.29 & 97.62 \\
ecoli-0-1-4-7\_vs\_5-6 & 87 & 76 & 67.32 & 90 & 90 & 91 & 94 & 94 \\
haber & 77.17 & 77.17 & 57.96 & 76.09 & 78.26 & 82.61 & 76.09 & 79.35 \\
haberman & 77.17 & 77.17 & 57.96 & 76.09 & 76.09 & 82.61 & 70.65 & 76.09 \\
haberman\_survival & 77.17 & 78.26 & 57.96 & 78.26 & 76.09 & 82.61 & 73.91 & 78.13 \\
heart-stat & 90.12 & 85.93 & 58.69 & 81.89 & 81.89 & 90.12 & 81.48 & 86.42 \\
led7digit-0-2-4-5-6-7-8-9\_vs\_1 & 92.23 & 78.35 & 66.77 & 94.74 & 94.74 & 93.23 & 84.96 & 94.74 \\
mammographic & 79.58 & 80.28 & 60.41 & 82.35 & 82.01 & 83.04 & 75.43 & 83.74 \\
monks\_2.csv & 67.84 & 71.29 & 68.42 & 65.87 & 70.89 & 72.23 & 69.43 & 72.43 \\
monks\_3 & 75.45 & 59.88 & 59.7 & 43.11 & 43.11 & 78.44 & 88.62 & 92.22 \\
musk\_1 & 68.53 & 52.66 & 59.15 & 81.44 & 84.62 & 72.03 & 79.02 & 79.72 \\
new-thyroid1 & 88.46 & 85.38 & 66 & 86 & 86 & 89.23 & 81.54 & 86.15 \\
oocytes\_merluccius\_nucleus\_4d & 64.82 & 63.19 & 59.58 & 82.74 & 80.71 & 69.06 & 75.9 & 81.13 \\
planning & 82.25 & 85.87 & 83.49 & 80.87 & 85.42 & 86.84 & 84.26 & 88.67 \\
spectf & 76.54 & 70.4 & 62.39 & 85.19 & 83.95 & 80.25 & 81.48 & 81.48 \\
tic\_tac\_toe & 75.69 & 76.88 & 68.66 & 96.65 & 96.65 & 99.65 & 99.65 & 98.61 \\
vehicle1 & 76.38 & 73.62 & 59.44 & 81.86 & 81.68 & 79.53 & 83.46 & 81.89 \\
vehicle2 & 71.65 & 53.54 & 64.86 & 90.03 & 90.85 & 94.09 & 93.31 & 92.52 \\
vertebral\_column\_2clases & 75.27 & 68.82 & 63.13 & 82.25 & 81.4 & 69.89 & 88.17 & 72.04 \\
wpbc & 77.97 & 57.63 & 60 & 70.97 & 69.49 & 77.97 & 77.97 & 71.19 \\
yeast-0-2-5-6\_vs\_3-7-8-9 & 83.71 & 64.9 & 66.23 & 89.71 & 89.71 & 88.08 & 62.91 & 90.07 \\
yeast-0-2-5-7-9\_vs\_3-6-8 & 86.79 & 68.55 & 66.67 & 95.68 & 95.35 & 87.5 & 64.24 & 95.03 \\
yeast-0-5-6-7-9\_vs\_4 & 81.19 & 56.6 & 68.29 & 76.1 & 93.71 & 84.91 & 76.1 & 94.34 \\
yeast-2\_vs\_4 & 85.81 & 54.19 & 67.7 & 95.48 & 96.77 & 87.74 & 87.1 & 83.23 \\
yeast3 & 79.91 & 79.03 & 67.15 & 85.07 & 86.72 & 80.04 & 86.77 & 89.46 \\ \hline
Average ACC & 80.71 &72.85 & 65.01 & 83.22 & 84.02 & 84.23 &  82.13	& \textbf{86.18}
 \\ \hline
Average Rank & 4.97& 6.02& 7.55 & 3.85& 3.76& 3.02& 4.42 & \textbf{2.42}
 \\ \hline
\multicolumn{9}{l}{$^{\dagger}$ represents the proposed models. Bold text denotes the model with the highest average ACC.}
\end{tabular}}
\end{table*}
\begin{table*}[ht!]
\centering
    \caption{Classification ACC over UCI and KEEL datasets with different percentages of label noise.}
    \label{Classification performance in label noise.}
    \resizebox{0.9\linewidth}{!}{
\begin{tabular}{lccccccccc}
\hline
Datasets & Noise & SVM \cite{cortes1995support} & GBSVM \cite{xia2022gbsvm} & TSVM \cite{khemchandani2007twin}  & RVFLwoDL  \cite{huang2006extreme} & RVFL \cite{pao1994learning} & GBTSVM \cite{quadir2024granularTSVM} & HF-GBTSVM & EF-GBTSVM$^{\dagger}$ \\ \hline
\multirow{4}{*}{chess\_krvkp} & $5 \%$ & 84.25 & 84.78 & 83.85 & 90.68 & 90.68 & 87.9 & 87.07 & 93.22 \\
 & $10 \%$ & 84.36 & 77.75 & 85.1 & 91.89 & 92.64 & 87.28 & 87.07 & 90.82 \\
 & $15 \%$ & 85.61 & 82.83 & 84.37 & 89.74 & 90.12 & 77.89 & 89.68 & 90.41 \\
 & $20 \%$ & 86.55 & 64.55 & 82.81 & 90.2 & 90.7 & 86.34 & 80.4 & 88.11 \\ \hline
\multirow{4}{*}{led7digit-0-2-4-5-6-7-8-9\_vs\_1} & $5 \%$ & 83.23 & 78.35 & 83.98 & 80.17 & 81.1 & 78.2 & 83.61 & 84.6 \\
 & $10 \%$ & 83.23 & 90.98 & 84.74 & 94.74 & 94.74 & 84.96 & 83.61 & 94.98 \\
 & $15 \%$ & 83.23 & 68.42 & 84.74 & 93.98 & 92.48 & 87.97 & 95.97 & 83.46 \\
 & $20 \%$ & 93.23 & 80.41 & 83.74 & 83.98 & 83.98 & 81.2 & 87.97 & 84.44 \\ \hline
\multirow{4}{*}{monks\_3} & $5 \%$ & 73.65 & 59.88 & 77.25 & 95.21 & 95.21 & 77.25 & 92.81 & 92.81 \\
 & $10 \%$ & 73.05 & 70.66 & 76.65 & 94.01 & 94.01 & 78.44 & 92.81 & 94.42 \\
 & $15 \%$ & 73.05 & 70.06 & 70.44 & 91.62 & 92.22 & 73.89 & 94.42 & 95.03 \\
 & $20 \%$ & 71.86 & 80 & 71.26 & 89.22 & 89.82 & 80.24 & 82.04 & 90.84 \\ \hline
\multirow{4}{*}{tic\_tac\_toe} & $5 \%$ & 75.35 & 70.47 & 95.65 & 97.65 & 97.65 & 99.65 & 98.28 & 99.65 \\
 & $10 \%$ & 76.04 & 70.97 & 89.65 & 92.31 & 92.31 & 99.65 & 98.28 & 99.65 \\
 & $15 \%$ & 74.65 & 60.76 & 89.65 & 97.65 & 98.31 & 99.65 & 97.65 & 98.61 \\
 & $20 \%$ & 73.96 & 63.33 & 89.65 & 92.92 & 97.92 & 99.65 & 93.75 & 98.96 \\ \hline
\multirow{4}{*}{yeast3} & $5 \%$ & 80.81 & 81.26 & 81.48 & 91.17 & 91.27 & 84.98 & 91.48 & 94.55 \\
 & $10 \%$ & 89.69 & 82.74 & 89.69 & 93.95 & 90.5 & 91.03 & 91.48 & 90.77 \\
 & $50 \%$ & 89.46 & 82.87 & 88.79 & 90.83 & 90.72 & 90.13 & 80.7 & 89.46 \\
 & $20 \%$ & 88.12 & 76.23 & 80.81 & 90.38 & 93.05 & 86.55 & 89.01 & 93.24 \\ \hline
 \multirow{4}{*}{Average ACC} & $5\%$ & 79.46 & 74.95 & 84.44 & 90.98 & 91.18 & 85.6 & 90.65 & \textbf{92.97} \\
& $10\%$ & 81.27 & 78.62 & 85.17 & 93.38 & 92.84 & 88.27  & 90.65 & \textbf{94.13} \\
& $15\%$ & 81.2 & 72.99 & 83.6 & 92.76 & \textbf{92.77} & 85.91  & 91.68 & 91.39 \\
& $20\%$ & 82.74 & 72.9 & 81.65 & 89.34 & 91.09 & 86.8  & 86.63 & \textbf{91.12} \\ \hline
\multicolumn{9}{l}{$^{\dagger}$ represents the proposed models. Bold text denotes the model with the highest average ACC.}
\end{tabular}}
\end{table*}


We now conduct statistical tests to determine the significance of the results. Specifically, we use the Friedman test \cite{demvsar2006statistical} to evaluate whether there are statistically significant differences between the models. Under the null hypothesis, it is assumed that all models have the same average rank, suggesting that they perform at the same level. The Friedman statistic, which follow the chi-squared distribution $(\chi_F^2)$ with $(q - 1)$ degrees of freedom (d.o.f), and its computation involves: $\chi_F^2 = \frac{12P}{q(q+1)}\left[ \sum_j \mathscr{R}_j^2 - \frac{q(q+1)^2}{4} \right]$. The $F_F$ statistic is computed as $F_F = \frac{(P-1)\chi_F^2}{P(q-1) - \chi_F^2}$, where the $F$-distribution possesses degrees of freedom $(q - 1)$ and $(P - 1) \times (q - 1)$. For $q=8$ and $P=32$, the obtained values are $\chi_F^2 = 85.230$ and $F_F = 19.0396$. The critical value $F_F(7, 196) = 2.0565$ at a $5\%$ level of significance. We reject the null hypothesis as $19.0396 > 2.0565$. Thus, there exists a statistically significant difference among the models being compared. Next, we employ the Nemenyi post hoc test to examine the pairwise differences between the models. The critical difference $(C.D.)$ value is calculated as $C.D. = q_{\alpha}\sqrt{\frac{q(q+1)}{6P}}$. The critical value $q_{\alpha} = 3.031$ is employed to evaluate $8$ models at a significance level of $5\%$. After a simple calculation, we obtained $C.D. = 1.8561$. The difference in average ranks between pairs of models EF-GBTSVM with SVM, GBSVM, TSVM, RVFLwoDL, RVFL, GBTSVM, and HF-GBTSVM are $2.55$, $3.60$, $5.13$, $1.43$, $1.34$, $0.60$, and $2.00$, respectively. The proposed EF-GBTSVM model exhibits significant differences from the baseline models, except RVFL, RVFLwoDL, GBTSVM, and HF-GBTSVM. However, the average rank of the proposed EF-GBTSVM model surpasses the RVFL, RVFLwoDL, GBTSVM, and HF-GBTSVM models. Hence, the proposed EF-GBTSVM model showcases superior performance against the baseline models and HF-GBTSVM. 
\begin{table*}[ht!]
\centering
    \caption{Comparison results on benchmark NDC datasets for the proposed EF-GBTSVM model with the baseline models.}
    \label{NDC}
    \resizebox{0.9\linewidth}{!}{
\begin{tabular}{lccccccccc}
\hline
NDC datasets & \begin{tabular}[c]{@{}c@{}}SVM \cite{cortes1995support}\\ ACC $(\%)$\\ Time (s)\end{tabular} & \begin{tabular}[c]{@{}c@{}}GBSVM \cite{xia2022gbsvm}\\ ACC $(\%)$\\ Time (s)\end{tabular} & \begin{tabular}[c]{@{}c@{}}TSVM \cite{khemchandani2007twin}\\ ACC $(\%)$\\ Time (s)\end{tabular} & \begin{tabular}[c]{@{}c@{}}RVFLwoDL \cite{huang2006extreme}\\ ACC $(\%)$\\ Time (s)\end{tabular} & \begin{tabular}[c]{@{}c@{}}RVFL \cite{pao1994learning}\\ ACC $(\%)$\\ Time (s)\end{tabular} & \begin{tabular}[c]{@{}c@{}}GBTSVM \cite{quadir2024granularTSVM} \\ ACC $(\%)$\\ Time (s)\end{tabular} & \begin{tabular}[c]{@{}c@{}}HF-GBTSVM\\ ACC $(\%)$\\ Time (s)\end{tabular} & \begin{tabular}[c]{@{}c@{}}EF-GBTSVM$^{\dagger}$\\ ACC $(\%)$\\ Time (s)\end{tabular} \\ \hline
Train 1l & a & b & a & 78.63 & 78.3 & \textbf{85.77} & 78.87 & 83.01 \\
 &  &  &  & 0.944 & 0.944 & 0.3562 & 0.86 & 0.844 \\
Train 3l & a & b & a & 79.52 & 80.97 & 80.52 & 80.1 & \textbf{81.33} \\
 &  &  &  & 2.702 & 2.702 & 0.906 & 2.127 & 2.015 \\
Train 5l & a & b & a & 81.15 & 82.38 & 81.12 & 81.67 & \textbf{83.97} \\
 &  &  &  & 11.454 & 11.454 & 1.5326 & 2.924 & 2.33 \\
Train 1m & a & b & a & 81.57 & 81.66 & 79.42 & 81.95 & \textbf{82.45} \\
 &  &  &  & 69.031 & 69.031 & 2.7863 & 6.244 & 6.25 \\
Train 3m & a & b & a & 80.44 & 80.87 & 78.04 & 82.04 & \textbf{82.22} \\
 &  &  &  & 76.725 & 76.725 & 8.1229 & 19.054 & 16.744 \\
Train 5m & a & b & a & 83.61 & 84.04 & 78.68 & \textbf{84.18} & 83.71 \\
 &  &  &  & 83.1 & 83.93 & 12.8752 & 35.982 & 30.312 \\
Train 1cr & a & b & a & 82.79 & 82.88 & 84.68 & 84.66 & \textbf{85.21} \\
 &  &  &  & 92.64 & 91.44 & 54.38 & 62.33 & 57.212 \\ \hline
 \multicolumn{7}{l}{\begin{tabular}[c]{@{}l@{}} $a$ Terminated because of out of memory. \\ $b$ Experiment is terminated because of the out of bound issue shown by PSO algorithm. \\ $^{\dagger}$ represents the proposed models. Bold text denotes the model with the highest ACC. \end{tabular}}
\end{tabular}}
\end{table*}

\begin{table*}[ht!]
\centering
    \caption{Comparison results on benchmark AD and SCZ datasets for the proposed EF-GBTSVM model with the baseline models.}
    \label{AD and Schizophrenia}
    \resizebox{1\linewidth}{!}{
\begin{tabular}{lccccc}
\hline
Dataset & GBSVM \cite{xia2022gbsvm}  & RVFL \cite{pao1994learning} & GBTSVM \cite{quadir2024granularTSVM} & HF-GBTSVM & EF-GBTSVM$^{\dagger}$ \\
 & (ACC $(\%)$, Specificity) &  (ACC $(\%)$, Specificity)  &  (ACC $(\%)$, Specificity)   & (ACC $(\%)$, Specificity) & (ACC $(\%)$, Specificity) \\
 & (Precision, Recall)  & (Precision, Recall)  & (Precision, Recall)  & (Precision, Recall)  & (Precision, Recall) \\ \hline  
CN\_vs\_AD  & $(70.34, 89.57)$  & $(88, 89.71)$ & $(85.6, 80.45)$ & $(85.6, 91.18)$ & $(88.89, 82.35)$ \\
 &  $(62.5, 72.48)$  & $(87.5, 85.96)$ & $(85.6, 80.49)$ & $(88.24, 78.95)$ & $(77.78, 83.68)$ \\
CN\_vs\_MCI &  $(66.44, 86.76)$  & $(72.87, 47.76)$ & $(69.57, 60.56)$ & $(65.43, 55.22)$ & $(73.28, 62.69)$ \\
  & $(83.64, 82.78)$  & $(75, 86.78)$ & $(74.57, 82.56)$ & $(74.14, 71.07)$ & $(78.63, 86.78)$ \\
MCI\_vs\_AD &  $(69.16, 49.25)$  & $(68.18, 88.7)$ & $(69.39, 89.44)$ & $(68.75, 92.17)$ & $(69.82, 90.43)$ \\
 &  $(75.54, 24.89)$  & $(58.06, 29.51)$ & $(60.14, 25.44)$ & $(62.5, 24.59)$ & $(59.26, 32.79)$ \\
Schizophrenia  & $(60.42, 77.39)$  & $(63.89, 77.89)$ & $(60, 82.89)$ & $(61.36, 40.91)$ & $(66.64, 86.36)$ \\
  & $(58.33, 78.26)$  & $(65.36, 79.89)$ & $(72.45, 81.42)$ & $(58.06, 80.91)$ & $(75, 81.82)$ \\ \hline
Average  & $(66.59, 75.7425)$   & $(73.235, 76.015)$ & $(71.14, 78.335)$ & $(70.285, 69.87)$ & $(\mathbf{74.6575}, \mathbf{80.4575})$ \\
  & $(64.6025, 64.6025)$  & $(70.535, 70.535)$ & $(73.19, 67.4775)$ & $(63.88, 63.88)$ & $(\mathbf{71.2675}, \mathbf{71.2675})$ \\ \hline
  \multicolumn{6}{l}{$^{\dagger}$ represents the proposed models. Bold text denotes the model with the highest average ACC.}
\end{tabular}}
\end{table*}

\subsection{Evaluation on UCI and KEEL Datasets with Label Noise}
While the UCI and KEEL datasets employed in our study are representative of real-world scenarios, it is crucial to acknowledge that the presence of impurities or noise in collected data can escalate due to various factors. In such circumstances, the development of a robust model becomes imperative, capable of effectively addressing and handling these challenging scenarios. To showcase the superiority of the proposed EF-GBTSVM model even in adverse conditions, the label noise is introduced at varying levels of $5\%$, $10\%$, $15\%$, and $20\%$. We have selected $6$ diverse UCI and KEEL datasets for our comparative analysis. The result presented in Table \ref{Classification performance in label noise.} demonstrates the effectiveness of these models compared to the baseline SVM, GBSVM, TSVM, RVFLwoDL, RVFL, GBTSVM, and HF-GBTSVM models. From the Table \ref{Classification performance in label noise.}, the proposed EF-GBTSVM model exhibited a top position compared to the existing models. The average ACC of the proposed EF-GBTSVM model at $5\%$, $10\%$, $15\%$, and $20\%$ level of noise are $92.97\%$, $94.13\%$, $91.39\%$ and $91.12\%$, respectively. The average ACC of the EF-GBTSVM model is highest at $5\%$, $10\%$, $15\%$, and $20\%$ level of noise except RVFL, RVFLwoDL, and HF-GBTSVM at $15\%$ level of noise. This significant difference in accuracies highlights the superior performance and effectiveness of the proposed EF-GBTSVM model relative to the existing baseline models. The observations mentioned above emphasize the importance of the proposed EF-GBTSVM model as a robust model with improved feature extraction capabilities. The proposed model showcases an adept ability to navigate and excel in scenarios where noise and impurities pose substantial challenges.
\subsection{Experiment on NDC Datasets}
The preceding comprehensive analyses consistently reveal the superior performance of the proposed EF-GBTSVM model in comparison to the baseline models across the majority of UCI and KEEL benchmark datasets. Now, we conduct an experiment utilizing the NDC datasets \cite{ndc} to emphasize the improved training speed and scalability of our proposed models. For this, the hyperparameters $d_1$ and $d_2$ are set to $1$ to reduce the training time. The sample sizes of these NDC datasets range from $1l$ to $1cr$, each containing $32$ features. The results depicted in Table \ref{NDC} demonstrate the efficiency and scalability of the proposed EF-GBTSVM model. Across the NDC datasets, our proposed model consistently surpass the baseline models in both ACC and training times, affirming their robustness and efficiency, especially when handling large-scale datasets. In terms of ACC, our proposed EF-GBTSVM model exhibits superior performance, achieving up to $3\%$ increase in ACC compared to baseline models on large datasets. Furthermore, our proposed model demonstrates reduced training time in comparison to the compared existing models. The experimental findings indicate that the proposed EF-GBTSVM model exhibits efficient training speed, surpassing some baseline models by several hundred or even a thousand times. This reduction in training time can be attributed to the considerably lower count of generated GBs on a dataset compared to the total number of samples.

\subsection{Evaluation on ADNI and Schizophrenia datasets}
To demonstrate the practical applicability of the proposed EF-GBTSVM model in real-world situations, we employ datasets from the Alzheimer's Disease Neuroimaging Initiative (ADNI) and Schizophrenia. To comprehensively assess the effectiveness of the proposed EF-GBTSVM model, we evaluate their performance using several metrics, such as ACC, Specificity, Precision, and Recall. 

Alzheimer’s disease (AD) is a progressive neurodegenerative disorder that severely impacts memory and cognitive functions. It often starts with mild cognitive impairment (MCI) as an early stage. It is important to note that not all patients with mild MCI go on to develop AD \cite{davatzikos2008individual}. Currently, the exact causes of AD are still not fully understood. However, accurate identification and diagnosis of AD are essential for delivering effective treatment, particularly during the early stages of the condition. Launched in 2003 by Michael W. Weiner, the ADNI project aims to explore various neuroimaging techniques, including magnetic resonance imaging (MRI), positron emission tomography (PET), and other diagnostic tools for AD, with a particular focus on the mild MCI stage. The feature extraction process used in this study adheres to the approach outlined in \cite{richhariya2021efficient}. The dataset includes three classification scenarios: CN versus AD (CN\_vs\_AD), CN versus MCI (CN\_vs\_MCI), and mild MCI versus AD (MCI\_vs\_AD). 

Schizophrenia (SCZ) is a severe mental disorder and one of the leading causes of disability worldwide. However, a substantial number of SCZ cases go untreated due to challenges like misdiagnosis, self-denial, and the social stigma associated with the condition. With the rise of social media, individuals living with SCZ are increasingly using these platforms to share their mental health struggles and explore support and treatment options. Additionally, machine learning techniques are increasingly applied to multimedia data and T1-weighted MRI datasets to diagnose SCZ. The data used in this study is sourced from the Center for Biomedical Research Excellence (COBRE)\footnote{\url{http://fcon_1000.projects.nitrc.org/indi/retro/cobre.html}}. The dataset consists of $74$ subjects in the healthy control group (average age $35.8 \pm 11.5$ years, ranging from $18$ to $65$ years) and $72$ subjects diagnosed with SCZ (average age $38.1 \pm 13.9$ years, ranging from $18$ to $65$ years). The CAT12 package\footnote{\url{http://www.neuro.uni-jena.de/cat/}}, implemented within the Statistical Parametric Mapping (SPM) toolbox version $12$\footnote{\url{https://www.fil.ion.ucl.ac.uk/spm/software/spm12/}}, is utilized for processing the images. The 3-D T1-weighted MRI scans were parcellated into white matter (WM), gray matter (GM), and cerebrospinal fluid, in addition to segmenting the skull, scalp, and air cavities. Using the high-dimensional diffeomorphic anatomical registration through the exponentiated Lie algebra algorithm (DARTEL), the gray matter (GM) images were normalized to the Montreal Neurological Institute (MNI) space. The smoothed gray matter (GM) images were created using a Gaussian kernel with an 8-mm full-width at half maximum (FWHM).

Table \ref{AD and Schizophrenia} presents the performance metrics for AD and Schizophrenia diagnosis, comparing the proposed EF-GBTSVM model with the baseline models. The proposed EF-GBTSVM model demonstrated the highest performance, achieving an average accuracy (ACC) of $74.6575\%$. The baseline RVFL model ranked second with an average ACC of $73.235\%$, followed by the compared GBTSVM model in third place with an average ACC of $71.14\%$. The HF-GBTSVM model came fourth, attaining an accuracy of $70.285\%$. For CN\_vs\_AD, EF-GBTSVM achieves the highest ACC ($88.89\%$) and specificity ($82.35\%$), outperforming all baseline models, particularly GBTSVM (($85.6\%$). The model’s balanced precision ($77.78\%$) and recall ($83.68\%$) demonstrate its superior ability to correctly identify AD patients while minimizing false positives. For CN\_vs\_MCI, our proposed EF-GBTSVM model shows superior ACC ($73.82\%$) and recall ($86.78\%$), outperforming GBSVM ($66.44\%$ ACC, $82.78\%$ Recall), and GBTSVM ($69.57\%$ ACC, $82.56\%$ Recall) on this challenging task. The model’s strong recall highlights its effectiveness in early-stage cognitive impairment detection, crucial for timely interventions. With an ACC of $69.82\%$ and a specificity of $90.43\%$, EF-GBTSVM exceeds the performance of all the baseline models. Although this task is inherently difficult, EF-GBTSVM’s high recall emphasizes its ability to correctly identify AD patients within the MCI population. For SCZ classification, the proposed EF-GBTSVM model achieved ACC of $66.64\%$, while the baseline models GBSVM, RVFL, and GBTSVM attained ACC of $60.42\%$, $63.89\%$, and $60.82\%$, respectively. This indicates that our proposed model demonstrates superior generalization capabilities compared to the baseline models. The specificity, precision, and recall of the proposed EF-GBTSVM model are $86.36\%$, $75\%$, and $81.82\%$, respectively. The proposed EF-GBTSVM model achieved the highest specificity, precision, and recall among the baseline models. Consequently, the proposed EF-GBTSVM consistently demonstrates exceptional performance by attaining high ACC across different scenarios, highlighting its superiority among the models. The overall results underscore the effectiveness of the proposed models in distinguishing between various cognitive states.

\subsection{Sensitivity Analyses}
To comprehensively understand the robustness of the proposed models, it is essential to analyze their sensitivity to hyperparameters. Therefore, we conduct the following sensitivity analyses to delve deeper into the behavior of the models:
\subsubsection{Sensitivity Analysis of Hyperparameters \texorpdfstring{$d_1$}{d1} and \texorpdfstring{$d_2$}{d2}}
\begin{figure*}[ht!]
\begin{minipage}{.23\linewidth}
\centering
\subfloat[aus]{\label{fig:1a}\includegraphics[scale=0.20]{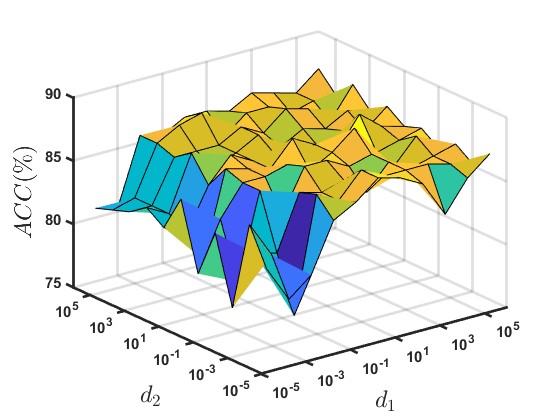}}
\end{minipage}
\begin{minipage}{.23\linewidth}
\centering
\subfloat[mammographic]{\label{fig:1b}\includegraphics[scale=0.20]{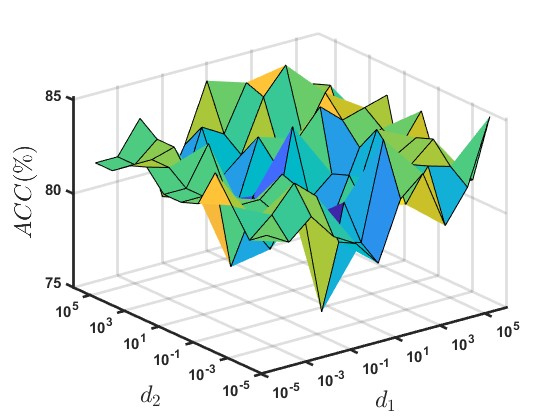}}
\end{minipage}
\begin{minipage}{.23\linewidth}
\centering
\subfloat[monks\_3]{\label{fig:1c}\includegraphics[scale=0.20]{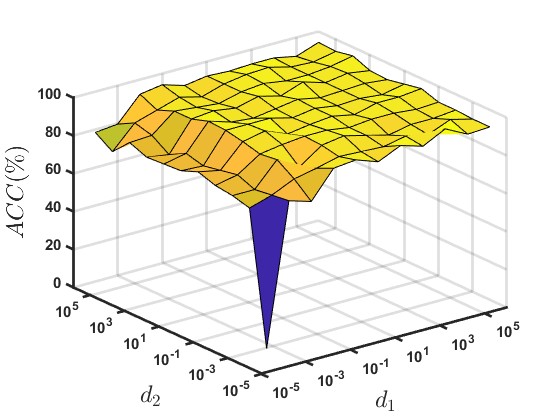}}
\end{minipage}
\begin{minipage}{.23\linewidth}
\centering
\subfloat[yeast-0-2-5-7-9-vs\_3-6-8]{\label{fig:1d}\includegraphics[scale=0.20]{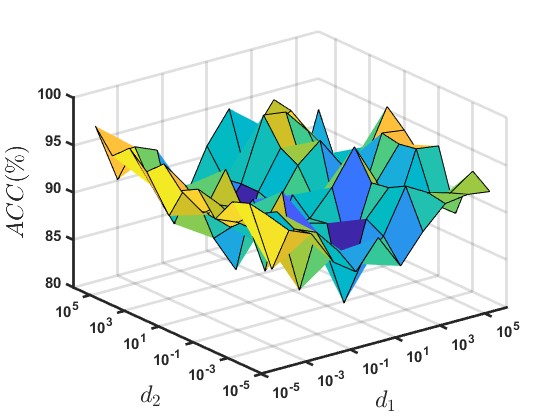}}
\end{minipage}
\caption{The effect of hyperparameter $(d_1, d_2)$ tuning on the accuracy (ACC) of some UCI and KEEL datasets on the performance of EF-GBTSVM.}
\label{The effect of hyperparameter c1 and c2}
\end{figure*}

To thoroughly grasp the nuanced effects of hyperparameters on the model's generalization capability, we systematically explore the hyperparameter space by varying the values of \( d_1 \) and \( d_2 \). This exploration allows us to identify configurations that maximize predictive accuracy and enhance the model's robustness to unseen data. The graphical representations in Fig \ref{The effect of hyperparameter c1 and c2} offer visual insights into how parameter tuning affects the ACC of our EF-GBTSVM model in the linear case. These visuals illustrate significant variations in model accuracy across different \( d_1 \) and \( d_2 \) values, underscoring the sensitivity of our model's performance to these hyperparameters. From Figs. \ref{fig:1a} and \ref{fig:1b}, the optimal performance of the proposed EF-GBTSVM model is observed within the $d_1$ and $d_2$ ranges of $10^{-1}$ to $10^{5}$ and $10^{-3}$ to $10^{5}$, respectively. From Figs. \ref{fig:1c} and \ref{fig:1d}, the ACC of the proposed EF-GBTSVM model archives the maximum when $d_1$ and $d_2$ ranges of $10^{-3}$ to $10^{5}$, respectively. Therefore, we recommend using $d_1$ and $d_2$ from the range $10^{-3}$ to $10^{5}$ for efficient results, although fine-tuning may be necessary depending on the dataset's characteristics for the proposed EF-GBTSVM model to achieve optimal generalization performance. 
\subsubsection{Effect of Parameter ``Act fun'' on the Performance of the Proposed EF-GBTSVM Model}
\begin{figure*}[ht!]
\begin{minipage}{.23\linewidth}
\centering
\subfloat[breast\_cancer]{\label{fig:2a}\includegraphics[scale=0.20]{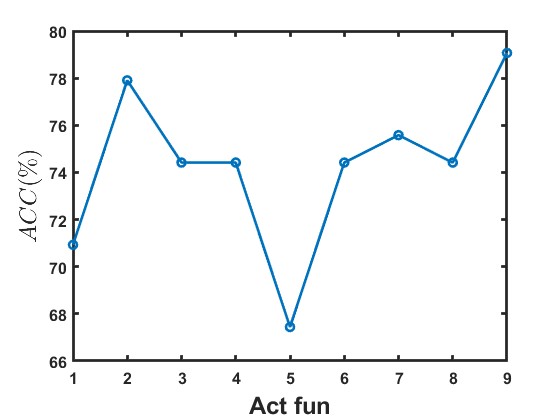}}
\end{minipage}
\begin{minipage}{.23\linewidth}
\centering
\subfloat[chess\_krvkp]{\label{fig:2b}\includegraphics[scale=0.20]{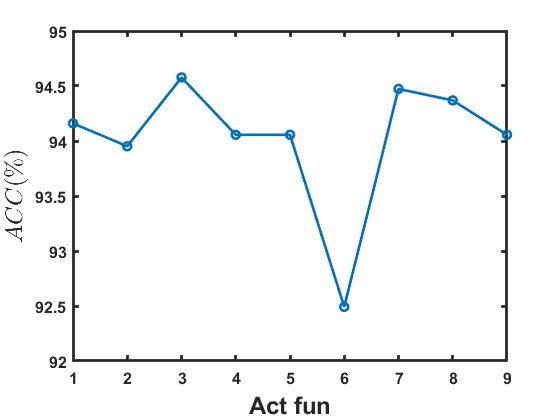}}
\end{minipage}
\begin{minipage}{.23\linewidth}
\centering
\subfloat[haber]{\label{fig:2c}\includegraphics[scale=0.20]{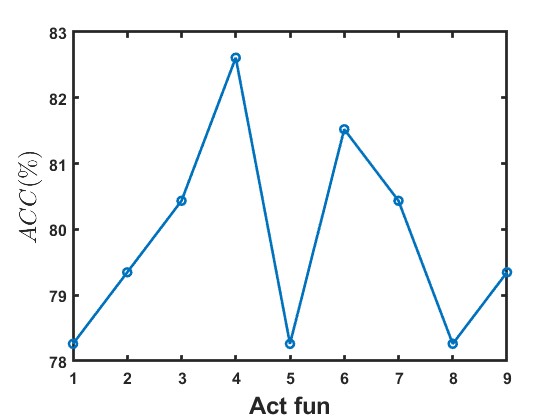}}
\end{minipage}
\begin{minipage}{.23\linewidth}
\centering
\subfloat[heart-stat]{\label{fig:2d}\includegraphics[scale=0.20]{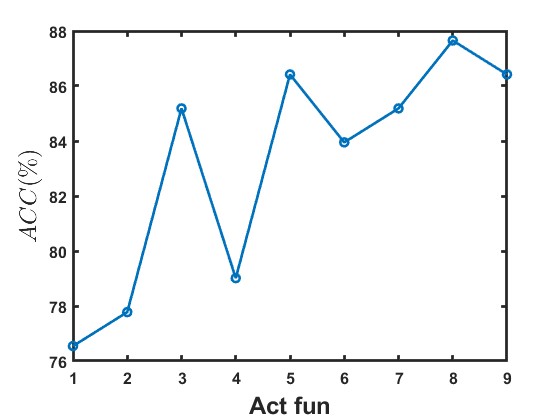}}
\end{minipage}
\caption{Effect of parameter ``Act fun'' on the performance of the proposed EF-GBTSVM model.}
\label{Effect of act}
\end{figure*}

The activation function significantly influences the performance of the EF-GBTSVM model. In our experiment, we tuned nine different activation functions. The indexing of these functions is as follows: 1) SELU, 2) ReLU, 3) Sigmoid, 4) Sine, 5) Hardlim, 6) Tribas, 7) Radbas, 8) Sign, and 9) Leaky ReLU. We investigate the relationship using Fig \ref{Effect of act} across datasets including breast\_cancer, chess\_krvkp, haber, and heart-stat. Our observation highlights a sensitivity to ``Act fun''. For instance, in the breast\_cancer dataset, activation function $9$ shows lower performance. Conversely, in the haber dataset, activation function $4$ exhibits superior performance. This variability underscores the mixed performance across different datasets with respect to Act fun, suggesting the importance of fine-tuning the activation function to optimize results effectively.
\begin{figure*}[ht!]
\begin{minipage}{.23\linewidth}
\centering
\subfloat[checkerboard\_Data]{\label{fig:2a1}\includegraphics[scale=0.20]{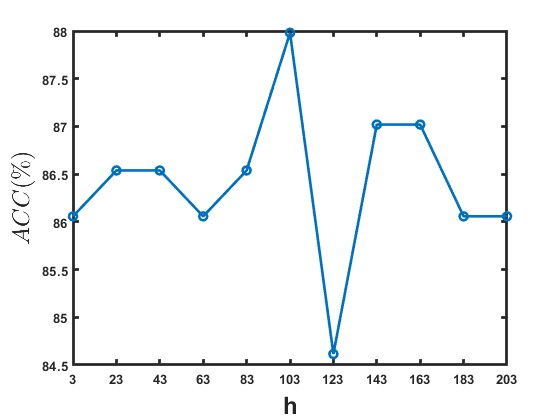}}
\end{minipage}
\begin{minipage}{.23\linewidth}
\centering
\subfloat[ecoli-0-1-4-6\_vs\_5]{\label{fig:2b1}\includegraphics[scale=0.20]{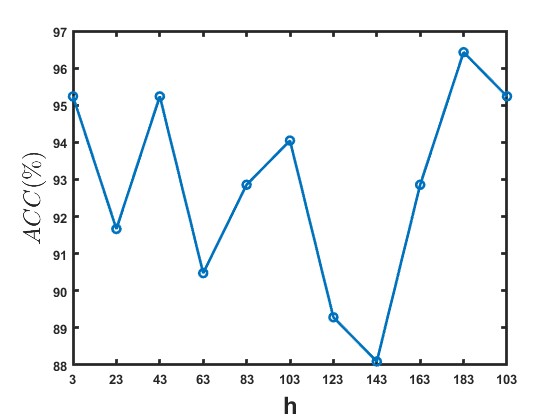}}
\end{minipage}
\begin{minipage}{.23\linewidth}
\centering
\subfloat[musk\_1]{\label{fig:2c1}\includegraphics[scale=0.20]{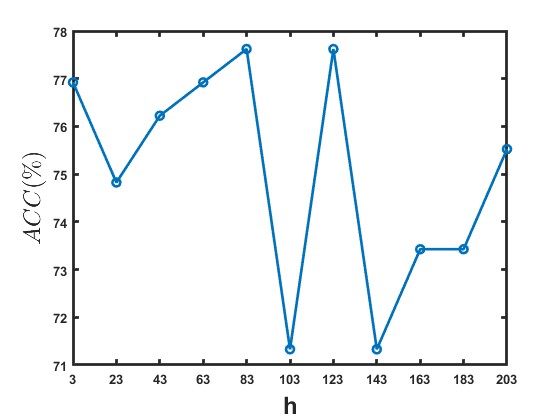}}
\end{minipage}
\begin{minipage}{.23\linewidth}
\centering
\subfloat[thyroid1]{\label{fig:2d1}\includegraphics[scale=0.20]{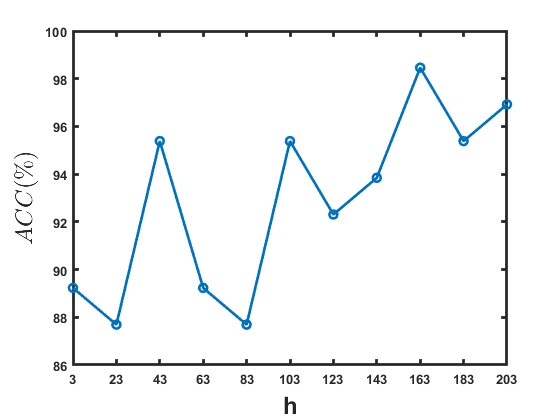}}
\end{minipage}
\caption{Effect of parameter $h$ on the performance of the proposed EF-GBTSVM model.}
\label{Effect of hidden node}
\end{figure*}
\subsubsection{Influence of the Numbers of Hidden Nodes \texorpdfstring{$h$}{h}}
The impact of hyperparameter $h$ (numbers of hidden nodes) is illustrated in Fig \ref{Effect of hidden node}. Our analysis reveals distinct trends for the EF-GBTSVM model. For thyroid dataset, the performance shows steady improvement with increasing $h$, plateauing at higher values like $h = 143$ or greater. Conversely, EF-GBTSVM  for checkerboard\_Data achieves peak performance at $h = 103$ and exhibits a gradual decline as $h$ increases beyond this point. In conclusion, our results underscore the impact of dataset characteristics on the performance of the proposed EF-GBTSVM model, stressing the importance of fine-tuning the parameter $h$ for optimal performance of the proposed EF-GBTSVM model.
\begin{table*}[ht!]
\centering
    \caption{Ablation study of the granular ball in the proposed EF-GBTSVM model compared with baseline TSVM, RVFLwoDL, and RVFL models over UCI and KEEL datasets.}
    \label{Classification performance in ablation Case.}
   \resizebox{0.9\linewidth}{!}{
\begin{tabular}{lccccccc}
\hline
Dataset    &  TSVM  \cite{khemchandani2007twin} & RVFLwoDL \cite{huang2006extreme} &  RVFL \cite{pao1994learning} & HF-TSVM  & HF-GBTSVM & EF-TSVM &  EF-GBTSVM \\ \hline
aus & 64.31 & 87.98 & 86.94 & 83.75 & 85.1 & 87.02 & 88.96 \\
checkerboard\_Data & 64.31 & 85.98 & 85.94 & 43.75 & 85.1 & 87.02 & 86.21 \\
chess\_krvkp & 67.41 & 90.2 & 90.41 & 69.04 & 90.09 & 97.5 & 98.16 \\
haber & 57.96 & 76.09 & 78.26 & 73 & 76.09 & 76.09 & 79.35 \\
monks\_3 & 59.7 & 43.11 & 43.11 & 82.608 & 88.62 & 93.41 & 92.22 \\ \hline
Average ACC  & 62.74  & 	76.67  & 76.93 &	70.43 &	85 &	88.21 &	88.98 \\ \hline
\end{tabular}}
\end{table*}
\subsection{Ablation Study}
We conducted an ablation study on the EF-GBTSVM model to confirm the importance of granular balls in improving model performance. In this study, we compared the EF-GBTSVM model's performance against baseline models, including TSVM, RVFLwoDL, and RVFL. Additionally, we trained HF-GBTSVM and EF-GBTSVM models using the original input samples instead of the generated granular ball centers, referring to these as HF-TSVM and EF-TSVM models, respectively. \\
From the results shown in Table \ref{Classification performance in ablation Case.}, the EF-GBTSVM model demonstrated superior performance across most datasets, achieving an average accuracy (ACC) of $88.98\%$, the highest among the models compared. The EF-TSVM model followed closely with an average ACC of $88.21\%$. This indicates that using granular balls, along with extracting features in the enhanced feature space, is crucial for enhancing model performance. \\
The ablation study highlights the significance of granular balls in the EF-GBTSVM model. By comparing the models trained with and without granular ball centers, we observed a clear performance advantage when incorporating granular balls. This suggests that the granular ball approach not only helps in capturing the underlying data structure but also improves the robustness and ACC of the model. The study underscores the critical role of both the granular ball framework and feature extraction in the enhanced feature space in achieving superior classification results.

\section{Conclusions}
\label{Conclusions}
This paper proposed an enhanced feature based granular ball twin support vector machine (EF-GBTSVM). EF-GBTSVM employs the coarse granularity of GBs as input, resulting in the formation of two non-parallel hyperplanes in the enhanced feature space. The proposed EF-GBTSVM reduces the impact of noise and outliers while alleviating the challenge of high computational costs. To demonstrate the effectiveness, robustness, scalability, and efficiency of our proposed EF-GBTSVM model, we conducted a comprehensive series of experiments supported by detailed statistical analyses. These analyses encompass various ranking schemes, including the Friedman and Nemenyi post hoc tests. Furthermore, by adding label noise to the UCI and KEEL datasets, we evaluated the robustness of our proposed model in comparison to baseline models under noisy conditions. We evaluated the performance of our models on NDC datasets, varying sample sizes from $1l$ to $1cr$ samples, with a specific emphasis on scalability. Our proposed models have showcased better efficiency, surpassing various baseline models by a factor of $100$ to $1000$. 
Furthermore, the superior performance of the proposed models on SCZ and AD shows their effectiveness in the real-world scenario. Although our proposed models have shown excellent performance in binary classification tasks, their effectiveness in multiclass scenarios has not yet been explored. A key direction for future research will be to adapt these models to effectively address multiclass problems. Additionally, enhancing the model by integrating a robust loss function \cite{sajid2024wavervfl}, leveraging the $L_1$ norm for regularization, and incorporating an intuitionistic fuzzy membership scheme could further bolster their robustness and efficacy.

\bibliographystyle{IEEEtranN}
\bibliography{refs.bib}

\end{document}